\documentclass[letterpaper, 10 pt, conference]{ieeeconf}  

\IEEEoverridecommandlockouts                              

\usepackage{amsmath} 
\usepackage{amssymb}  
\let\labelindent\relax
\usepackage[inline]{enumitem}
\usepackage{hyperref}
\usepackage{xcolor} 
\usepackage{array}
\usepackage{graphicx} 
\usepackage{booktabs}  
\usepackage{caption}   
\usepackage{lipsum} 
\usepackage{multirow}
\usepackage{booktabs} 
\usepackage{subcaption}
\usepackage{hyperref}
\hypersetup{
    colorlinks=true,
    linkcolor=blue,
    filecolor=magenta,      
    urlcolor=cyan,
    pdfpagemode=FullScreen,
    }

\usepackage{tcolorbox}

\usepackage{amsthm}

\newtheorem{criterion}{Criterion}

\usepackage{graphicx}
\usepackage{subcaption}
\usepackage{caption}
\usepackage{adjustbox}
\usepackage{lipsum}

\usepackage{adjustbox}
\usepackage{subcaption}
\usepackage{caption}
\title{\LARGE \bf
  Coverage Path Planning for Holonomic UAVs via Uniaxial-Feasible, Gap-Severity Guided Decomposition
}

\author{Pedro Antonio Alarcon Granadeno$^{1}$ and Jane Cleland-Huang$^{2}$
\thanks{*This work was supported by the USA National Science Foundation under Grant 1931962}
\thanks{$^{1}$Pedro Antonio Alarcon Granadeno is a PhD student in the Computer Science and Engineering department at the University of Notre Dame, IN, USA
        {\tt\small palarcon@nd.edu}}%
\thanks{$^{2}$Jane Cleland-Huang is a Professor in the Department of Computer Science and Engineering, University of Notre Dame, IN, USA     {\tt\small JaneHuang@nd.edu}}%
\thanks{Website and open-source  code found at \href{https://sites.google.com/view/cpp-holonomic-uav}{https://sites.google.com/view/cpp-holonomic-uav}}
}

\begin{document}
\maketitle
\thispagestyle{empty}
\pagestyle{empty}

\begin{abstract}
Modern coverage path planning (CPP) for holonomic UAVs in emergency response must contend with diverse environments where regions of interest (ROIs) often take the form of highly irregular polygons, characterized by asymmetric shapes, dense clusters of concavities, and multiple internal holes. Modern CPP pipelines typically rely on decomposition strategies that overfragment such polygons into numerous subregions. This increases the number of sweep segments and connectors, which in turn adds inter-region travel and forces more frequent reorientation. These effects ultimately result in longer completion times and degraded trajectory quality. We address this with a decomposition strategy that applies a recursive dual-axis monotonicity criterion, with cuts guided by a cumulative gap severity metric. This approach distributes clusters of concavities more evenly across subregions and produces a minimal set of partitions that remain sweepable under a parallel-track maneuver. We pair this with a global optimizer that jointly selects sweep paths and inter-partition transitions to minimize total path length, transition overhead, and turn count. We demonstrate that our proposed approach achieves the lowest mean path-length and completion-time overhead among 15 other CPP pipelines.
\end{abstract}

\section{Introduction}
Coverage Path Planning (CPP) generates trajectories that fully observe a Region of Interest (ROI) with minimal cost, typically measured in path length, turns, execution time, or energy use. A dominant strategy in aerial and ground robotics is parallel-track coverage, where back-and-forth sweeps (e.g., boustrophedon) tile the region. For convex ROIs, selecting an optimal sweep direction is straightforward. Huang \cite{huang2001optimal} showed that the direction perpendicular to the polygon’s minimal altitude minimizes turns, a widely accepted surrogate for maneuvering effort, execution time, and energy consumption \cite{hoffmann2024optimal}.

Real-world ROIs, however, are rarely convex. This leads to the need for polygon decomposition, where the region is subdivided into simpler subregions (often convex or monotone in a chosen direction), each assigned a local sweep pattern. Early strategies such as trapezoidal decomposition \cite{SEIDEL199151} and boustrophedon cellular decomposition (BCD) \cite{choset2000coverage} provided simple and topologically sound methods for this. Trapezoidal decomposition projects vertical rays from vertices to the boundary, yielding convex but often overfragmented partitions when edges are nearly parallel or reflex vertices are dense. BCD improves upon this by introducing a sweep-line that only places cuts at connectivity events (openings, closures, splits, merges) which yields fewer and more meaningful partitions. 
Both methods rely on a fixed global sweep direction, which limits flexibility and path efficiency \cite{bahnemann2021revisiting}.

\begin{figure}[!t]
    \centering
    \begin{subfigure}[b]{0.44\columnwidth}
        \centering
        \setlength{\fboxsep}{0pt}    
        \setlength{\fboxrule}{0.05pt} 
        \fbox{\includegraphics[width=\linewidth,clip,trim={20 0 76 0}]{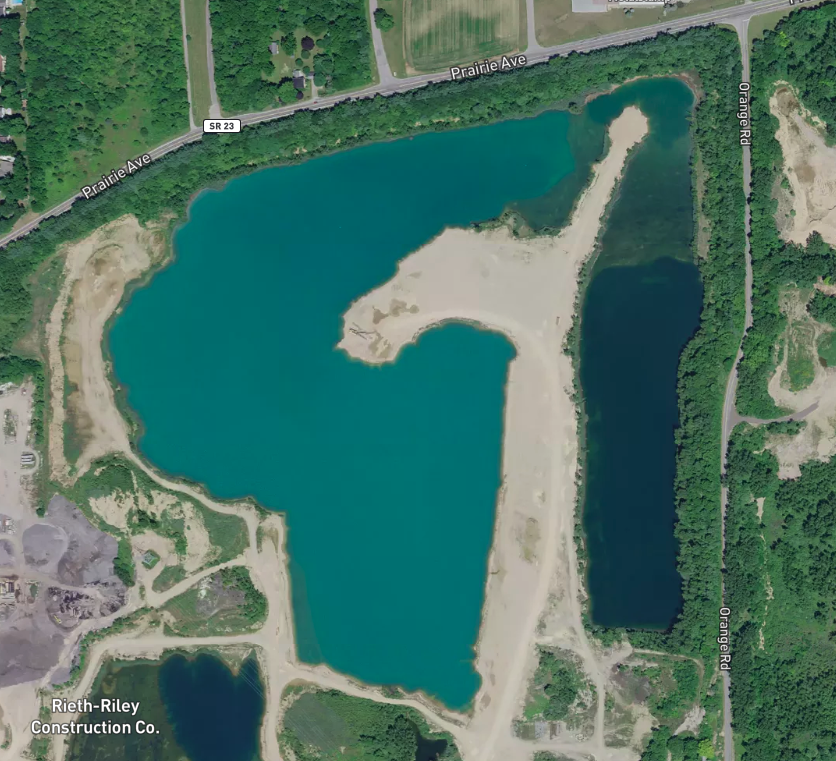}}
    \end{subfigure}
    \begin{subfigure}[b]{0.54\columnwidth}
        \centering
        \includegraphics[width=\linewidth,clip,trim={40 40 80 15}]{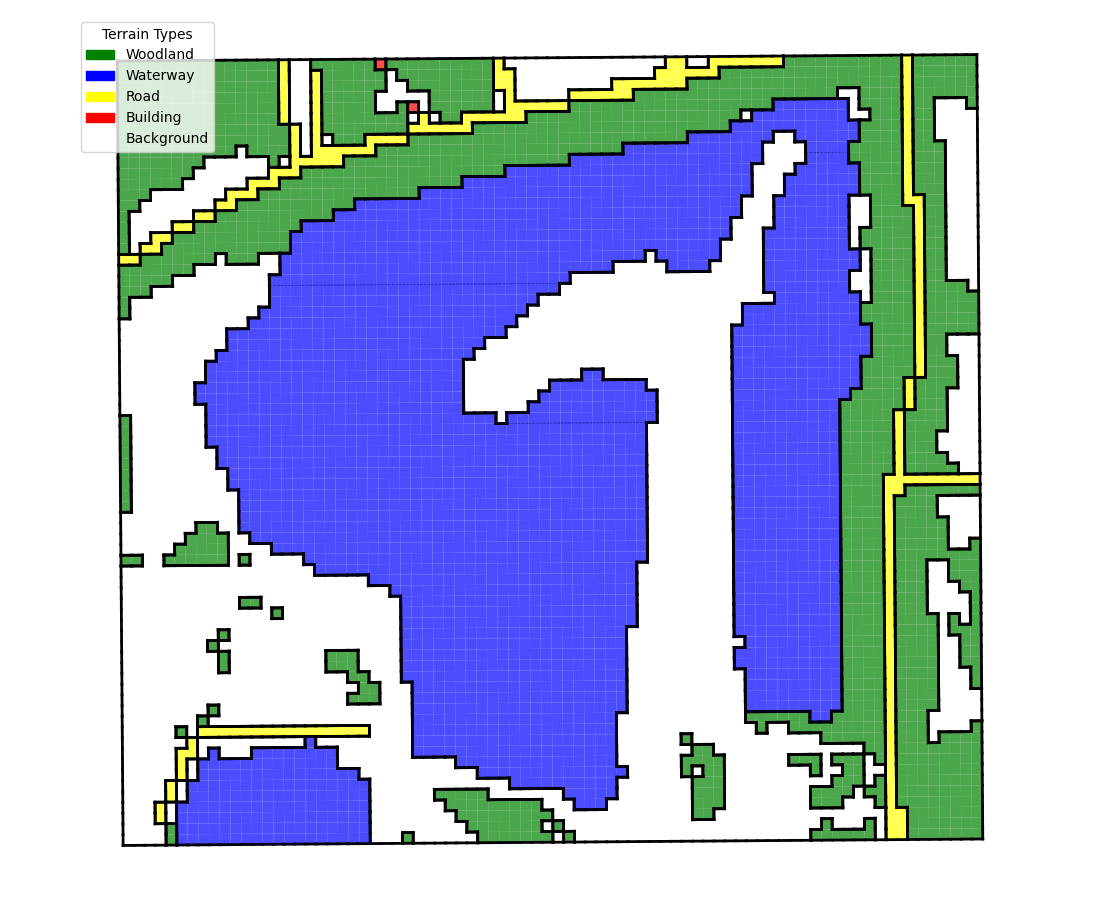}

    \end{subfigure}
\caption{\textbf{Left:} Raw satellite image. 
\textbf{Right:} Region-of-interest polygons derived via landcover segmentation and DFS-based component tracing. The extracted polygons vary widely in size and exhibit irregular, asymmetric, and elongated geometries typical of challenging coverage scenarios. }
    \label{fig:scene_reconstructed}
    \vspace{-12pt}
\end{figure}

Several contemporary CPP methods build upon these foundations but aim to improve adaptability and overall efficiency. Three recurring themes characterize much of this recent work. First, many algorithms perform decomposition across multiple sweep orientations rather than committing to a single axis from the outset. This is often paired with width-based surrogates, such as minimal altitude, to estimate turn cost \cite{hoffmann2024optimal}. Subregions are split to reduce cumulative altitude and, by extension, the number of passes \cite{LI2011876, yu2015coverage}. Second, several approaches introduce post-decomposition merging strategies to attenuate overfragmentation inefficiencies. Some rely on simple heuristics that merge adjacent partitions sharing the same sweep direction \cite{LI2011876, yu2015coverage}, while others apply optimization routines that merge subpolygons when the combined shape yields a more optimal width cost\cite{ huang2001optimal, nielsen2019convex, bochkarev2016minimizing}. Third, a subset of approaches incorporate inter-partition connectivity directly into the planning objective. These methods generate multiple candidate paths per region and solve a global sequencing problem (often modeled as a TSP or GTSP) to jointly minimize coverage and transition costs \cite{LI2011876, TORRES2016441, bochkarev2016minimizing}.

These strategies have shown strong performance on concave regions that are geometrically simple, such as polygons with rectangular outlines, moderate symmetry, or limited branching. These types of geometries commonly appear in synthetic benchmarks or agricultural settings where the spatial layout remains relatively uniform. However, this assumption fails in real-world aerial scenarios, particularly in applications like emergency response. Here, the ROI is often derived from landcover annotations, which are semantic segmentations that tightly trace the contours of natural and man-made features (Fig. \ref{fig:scene_reconstructed}). The resulting polygons are highly asymmetric, with winding boundaries, branched extensions, holes, and dense clusters of concavities. Classical and modern decomposition pipelines often overfragment these inputs, producing small partitions that yield abrupt changes in sweep alignment and excessive inter-region connectors downstream. Several core issues drive this overfragmentation: (i) \textbf{width-guided decomposition} (e.g., \cite{LI2011876, yu2015coverage}) introduces competing objectives, where minimizing per-region turn cost often comes at the expense of a larger, more fragmented partition set; (ii) \textbf{fixed global sweep axes} (as in \cite{SEIDEL199151,choset2000coverage}) misalign with local geometry, forcing unnecessary cuts; and (iii) \textbf{post-hoc merging is tightly constrained}, typically allowed only when the merged region remains convex and yields a strictly better width cost (as in \cite{LI2011876,huang2001optimal,bochkarev2016minimizing,nielsen2019convex}).

To address these limitations, we introduce a recursive decomposition strategy built around a relaxed geometric criterion. A region is considered sweepable if it is monotone with respect to at least one cardinal axis, a property we refer to as uniaxial feasibility. This criterion replaces stricter assumptions such as convexity, global sweep alignment, or width minimization, and is used both to determine when a region must be split and whether adjacent regions can be merged. When a region violates feasibility, we apply a recursive partitioning process guided by a cumulative gap severity metric that identifies and distributes clusters of concavity. Unlike prior methods that prioritize turn cost or axis alignment during decomposition, we aim to distribute regions of discontinuity evenly across resulting partitions. This yields a smaller partition set composed of larger, sweepable regions, leading to lower global path length and execution time as compared to contemporary CPP pipelines.

\section{Related Work}
\label{sec:related-work}

Decomposition is a core strategy in modern Coverage Path Planning (CPP) for simplifying coverage over complex regions. Recent CPP surveys note the recurring use of polygonal width as a practical surrogate for turn-related maneuvering cost \cite{hoffmann2024optimal}. Measured as the shortest distance between parallel supporting lines orthogonal to a candidate sweep direction, width affects the number of parallel coverage passes (for fixed sweep spacing) and thus lane transitions and turns, making it useful for decomposition, sweep-direction selection, and merging decisions. One influential early example is Huang's Multi-Strip Approximation (MSA) \cite{huang2001optimal}, which extends classical Boustrophedon decomposition. MSA sweeps orthogonal to each polygon edge, places dividers at topological changes, overlays them to form monotone partitions, and uses dynamic programming to evaluate merges and assign sweep directions that minimize total width.

Later approaches introduced width surrogates at different stages of the planning pipeline. Li et al. \cite{LI2011876} use a greedy recursive strategy that places axis-aligned cuts at reflex vertices, scores each candidate by the cumulative width of the resulting sub-polygons, and selects the minimum-cost cut until all partitions are convex. Yu and Hung \cite{yu2015coverage} instead extend trapezoidal decomposition with four event types (floor/ceiling, convex/concave) and apply it across multiple orientations orthogonal to polygon or obstacle edges, selecting the convex decomposition with the lowest total width. Both methods optionally merge adjacent convex regions that share a common sweep direction. Other methods postpone width optimization to the merging stage. Nielsen et al. \cite{nielsen2019convex} generate convex partitions by extending incident edges at reflex vertices until they intersect the boundary, then enumerate all valid convex unions and compute their widths. An integer linear program selects a subset of merged regions that together minimize total width. Bochkarev and Smith \cite{bochkarev2016minimizing} start from an existing convex decomposition and iteratively refine it. At each reflex vertex, they define a feasible cut space within the visible cone formed by adjacent edges, sample chords, and evaluate each by the sum of minimum altitudes. Cuts that lower total width replace previous ones, and the process continues until convergence.

Other approaches emphasize global optimization over existing partitions instead of decomposition. Torres et al. \cite{TORRES2016441} define four path alternatives per region, combining sweep orientation and traversal direction to determine entry and exit points. A global planner selects visitation order and local configuration to minimize total path length, jointly considering in-region coverage and inter-region transitions, while heuristically pruning the solution space. Jimenez et al. \cite{jimenez2007optimal} take a genetic approach, encoding the entire plan as a chromosome. The head specifies region order and entry/exit configurations, while the body selects local paths from a template library. A fitness function combining travel and estimated turn cost guides the evolution of efficient plans.

Despite their effectiveness on well-structured regions, width-guided strategies struggle on irregular geometries due to competing objectives. Cuts that reduce turn cost often increase partition count, sweep-axis churn, and connector complexity. Post-hoc merging offers limited relief but is highly sensitive to early partitioning decisions and is typically constrained by stringent admissibility tests that accept unions only when surrogate width metrics improve. In contrast, our method avoids width-based optimization during decomposition and merging. We introduce a cumulative gap severity metric that evaluates sweep inefficiencies along both axes and guides recursive cuts to spread clustered concavities across subregions. A dual-axis monotonicity criterion governs both decomposition and merging: a region is preserved if it is monotone along at least one cardinal axis, and is split only when both fail. This flexibility retains complex but sweepable regions that traditional convex or width-driven methods would prematurely divide. Sweep direction is deferred to a final stitching phase that jointly optimizes path length, turn cost, and inter-region transitions across all partitions.
\section{CPP Problem Formulation}\label{sec:problem-formulation}

Coverage path planning (CPP) seeks a trajectory that enables a robot to observe every part of a planar region of interest (ROI), denoted $\widetilde{R} \subset \mathbb{R}^2$. Unlike ground-based robots that must remain within strict workspace boundaries due to physical constraints, this work considers a holonomic multirotor UAV, capable of concurrent linear and rotational motion at any planar heading. In our holonomic setting, heading is not a constraint on planar motion. The UAV may translate laterally and transition between successive path segments without first stopping to reorient its body. Consequently, corner transitions do not impose curvature or minimum turning radius constraints. Instead, the dominant time cost at segment changes arises from translational acceleration and deceleration limits, rather than from nonholonomic turning-radius constraints or higher-order continuity requirements such as bounded curvature ($k_{\max} = 1/r_{\min}$) or $C^3$ path smoothness assumptions as in~\cite{5461496}.

The UAV is equipped with a nadir-pointing camera with an effective symmetric field-of-view (FOV) half-angle $\phi/2$ at altitude $h$. Under a flat-ground, pinhole-camera model, the cross-track footprint width becomes $w=2h\tan (\phi/2)$. We adopt an approximate cellular decomposition setting~\cite{Choset2001Coverage}, where the ROI is modeled as a bounded polygonal region $\widetilde{R}$ that may contain holes. Let $\mathcal{O} = \{O_1, O_2, \dots, O_K\}$ denote the set of polygonal holes, each representing an obstacle or exclusion zone. A uniform grid of square cells with resolution equal to the UAV's footprint ($w \times w$) is overlaid onto $\widetilde{R}$. All cells that intersect $\widetilde{R}$ but do not lie entirely within any hole $O_k \in \mathcal{O}$ are retained. The union of these retained cells defines an expanded rectilinear region $R$. By construction, $R \supseteq \widetilde{R}$, which guarantees full coverage albeit at the cost of some negligible excess area near the boundary.

A region is considered fully covered once every retained cell in $R$ has been observed. A cell is marked as covered when the cumulative camera footprint intersects it beyond a threshold ratio $\alpha$ (typically close to 1). This defines the core principle of complete coverage in CPP. Additional objectives usually include minimizing path length ($L$), execution time ($T$), and energy use ($E$). In emergency response scenarios, which motivate this work, fast execution is essential. Victim survival rates decline sharply with time; for instance, 93.6\% of live extrications after earthquakes occur within the first 24 hours~\cite{RomKelman2020}. Accordingly, after guaranteeing complete coverage, this study adopts execution time as the \textit{primary evaluation metric} when comparing coverage strategies.

\section{Coverage Path Planning Algorithm} \label{sec:approach}

\begin{figure*}[t]
\centering

\begin{subfigure}[t]{.24\textwidth}
\centering
\includegraphics[width=\linewidth]{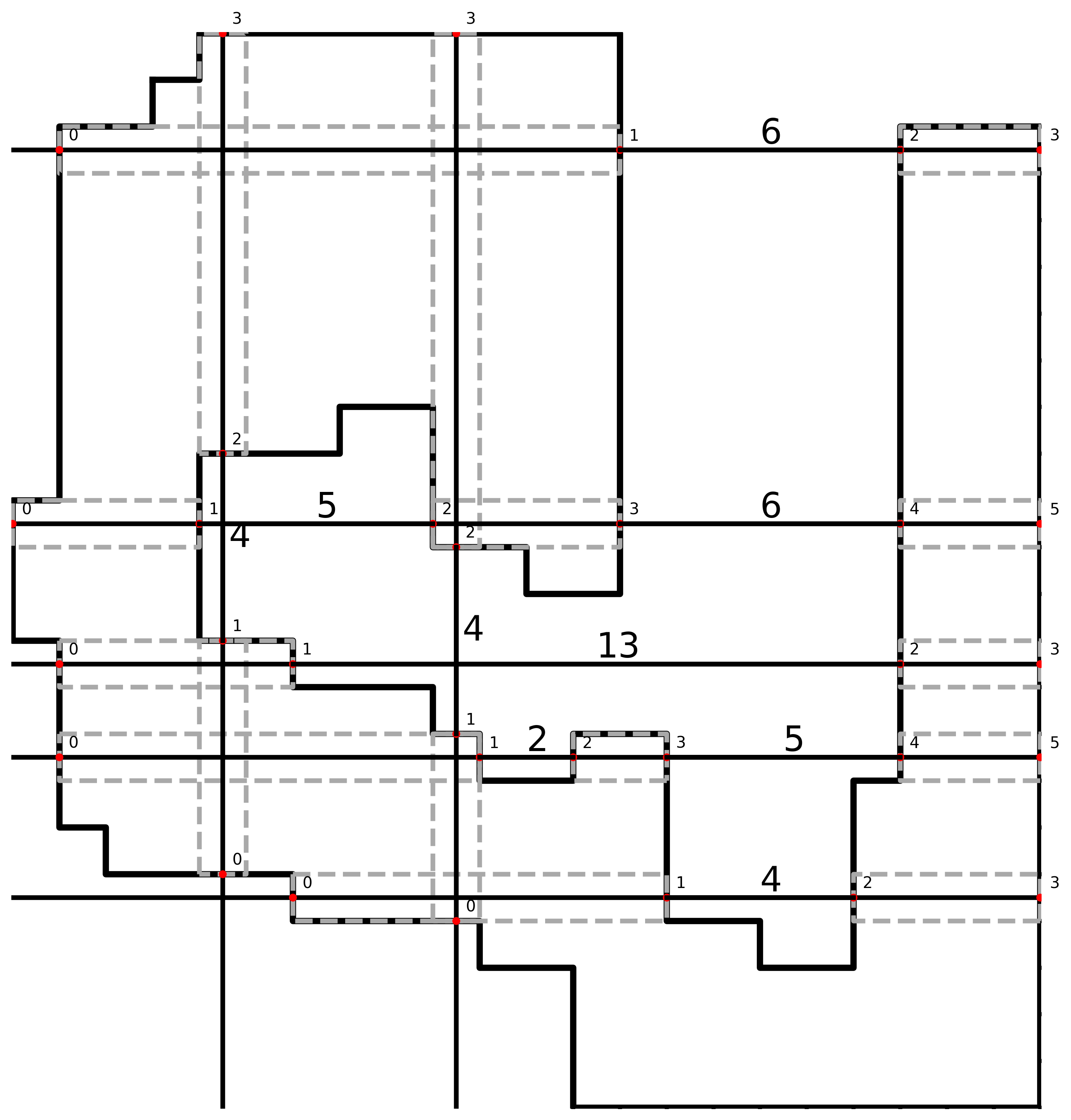}
\caption{}
\end{subfigure}%
\begin{subfigure}[t]{.24\textwidth}
\centering
\includegraphics[width=\linewidth]{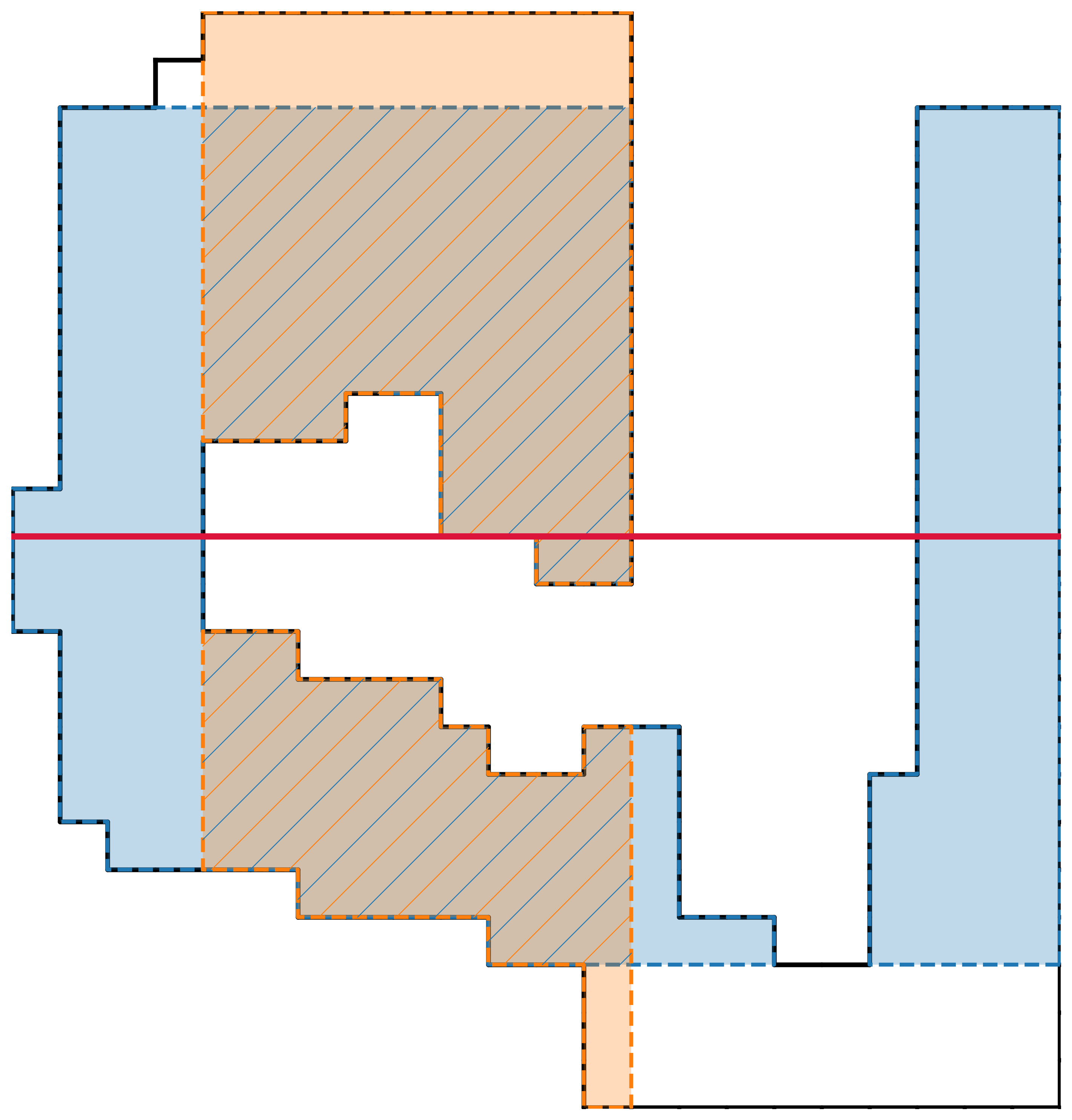}
\caption{}
\end{subfigure}%
\begin{subfigure}[t]{.24\textwidth}
\centering
\includegraphics[width=\linewidth]{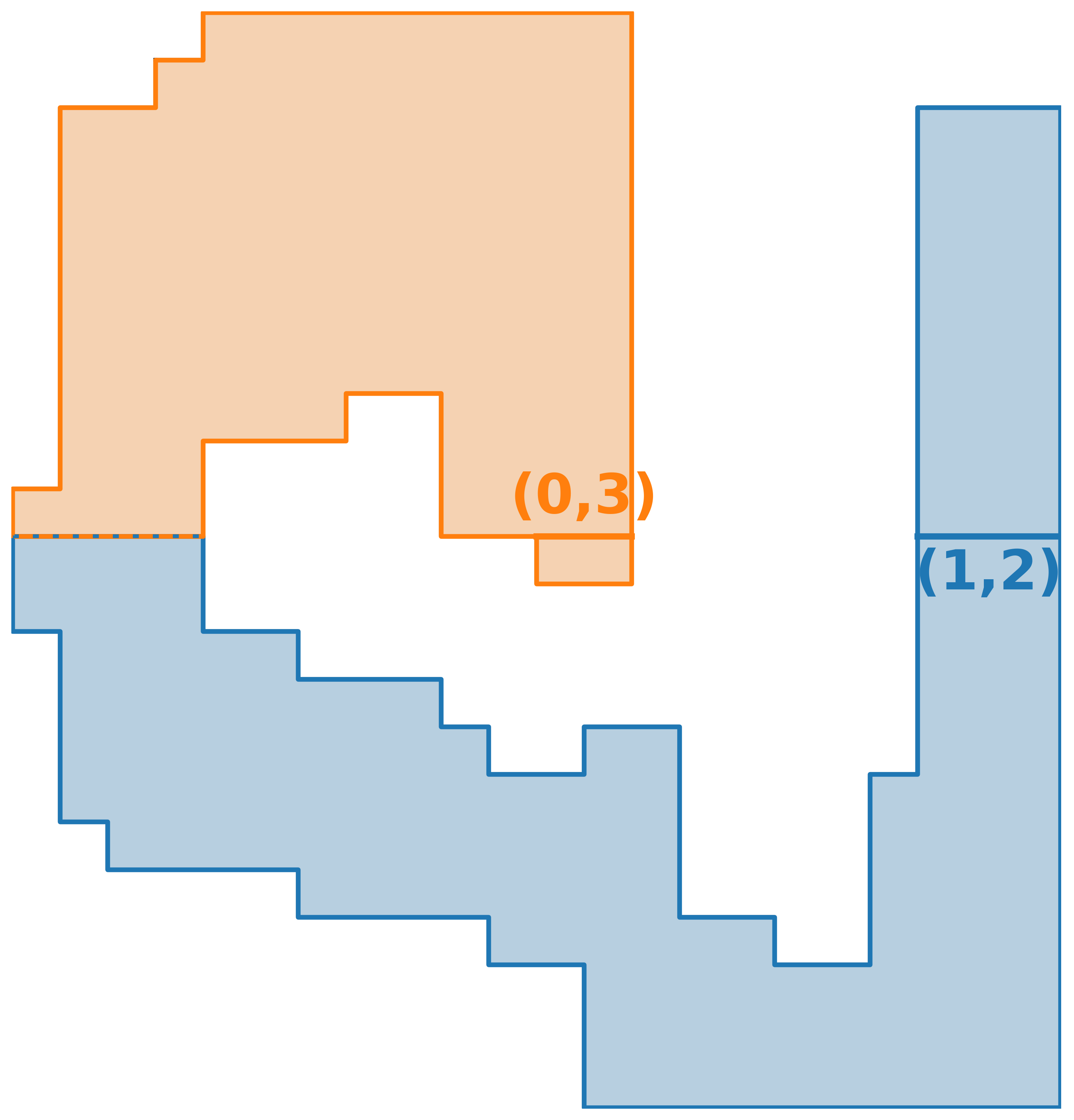}
\caption{}
\end{subfigure}%
\begin{subfigure}[t]{.24\textwidth}
\centering
\includegraphics[width=\linewidth]{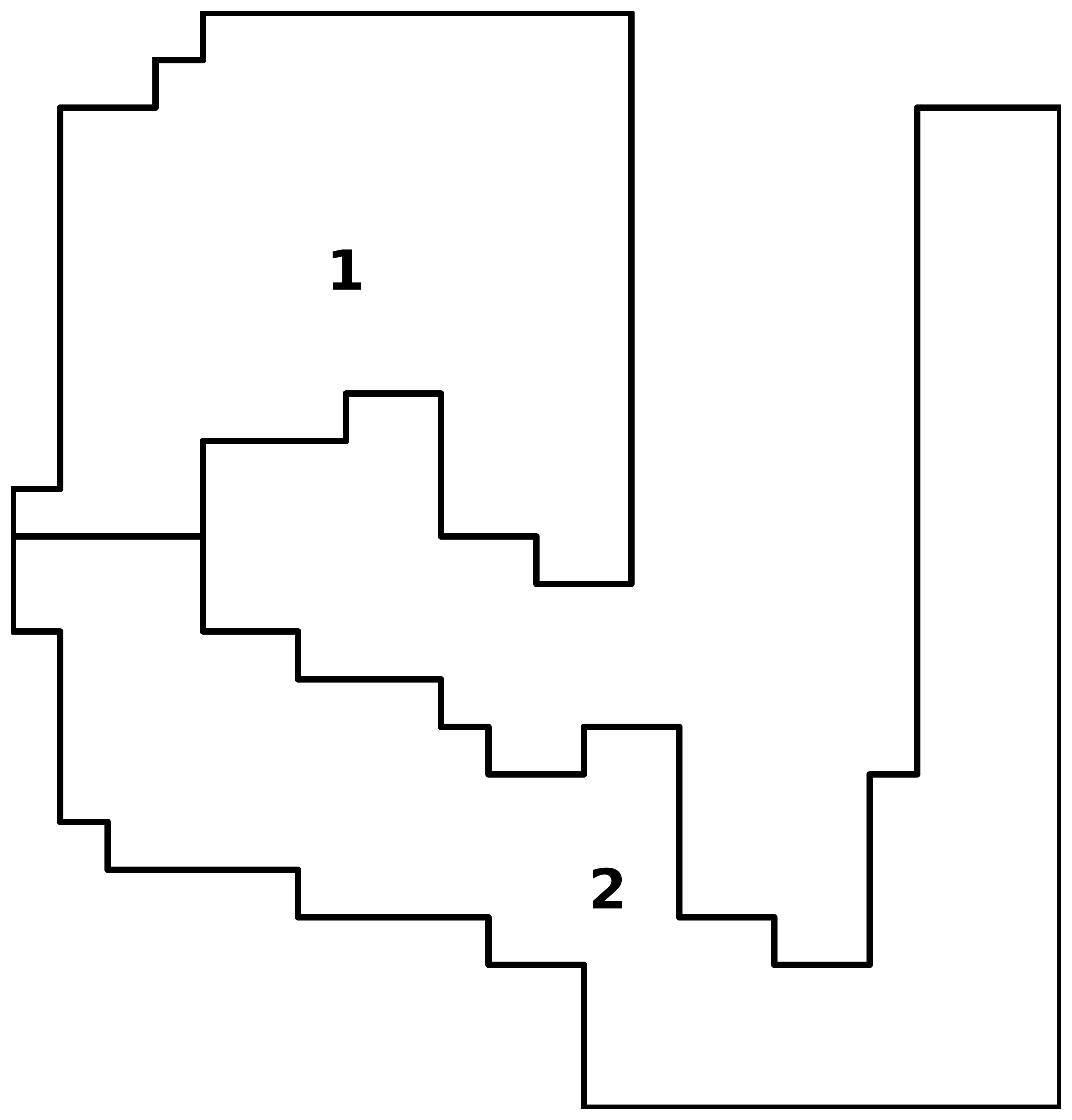}
\caption{}
\end{subfigure}

\caption{Overview of decomposition workflow on polygon \(P_{12}\). 
(a) Biaxial line sweep probes identifies and quantifies non-monotonic regions severities. 
(b) Cumulative regions of discontinuity (gap bands) guide partition placement. 
(c) Adjacent partitions are merged if their union satisfies the feasibility criterion. Merge candidates are labeled by partition indices, e.g., \((0,3)\) and \((1,2)\) denote proposed merges between partitions 0 and 3, and 1 and 2
(d) Final partitions are stored for downstream coverage planning.}
\label{fig:cpp-1x4-simple}
\end{figure*}

\subsection{Uniaxial-Feasible, Gap-Severity Guided Decomposition}

We operate on a rectilinear polygonal region \( P \subset \mathbb{R}^2 \), defined over a uniform grid of square cells with resolution equal to the UAV's camera footprint width 
$w$. The region may contain interior holes, which define non-observable zones. Our objective is to determine where $P$ can be swept directly using a parallel-track maneuver, or whether we must decompose into simpler subregions. To do so, we apply a recursive, line-sweep-inspired strategy that partitions a polygon only when it violates dual-axis monotonicity. Cut placements are guided by a cumulative \emph{gap severity} metric that quantifies how inefficient a traversal without decomposition would be inside regions of discontinuity. 

Many decomposition pipelines rely on stringent conditions: they fix a global sweep direction a priori (e.g., TD, BCD), decompose whenever a region is non-convex (e.g., Li), and merge only when convexity is preserved and width cost improves (e.g., Li, Huang). These constraints fragment regions that could otherwise be swept directly. To avoid premature cuts and preserve larger sweepable regions, we instead adopt a more permissive geometric rule called the uniaxial feasibility criterion, defined as follows:

\begin{criterion}[Uniaxial Feasibility Test]
\label{def:acceptability}
A polygon \( P \) is \textbf{acceptable} if it is monotone with respect to at least one of the two cardinal axes (horizontal or vertical). This criterion governs both decomposition and merging, that is, a region is decomposed only if it violates this condition, and adjacent regions are merged only if their union satisfies it.
\end{criterion}

This condition retains many concave shapes that would be discarded under convexity or width-based tests, while still guaranteeing parallel-track coverage. To evaluate axis monotonicity, we construct a finite family of axis-aligned sweep lines over the polygon boundary. Note that a rectilinear polygon is composed of individual grid cells of width \(w\), corresponding to the UAV's footprint as defined in Section \ref{sec:problem-formulation}. Let \(\partial P\) denote the polygon boundary. We extract the unique set of \(x\)- and \(y\)-coordinates from all cell vertices that lie on \(\partial P\), and sort them in increasing order:
\begin{equation}
X = \{x_1, x_2, \dots, x_n\}, \quad Y = \{y_1, y_2, \dots, y_m\}
\label{eq:XYcoords}
\end{equation}

Candidate sweep lines are then placed at midpoints between consecutive values. For the horizontal and vertical directions, these lines are:
\[
L^H_i = \left\{ y = \frac{y_i + y_{i+1}}{2} \right\}, \quad i = 1, \dots, m - 1,
\]
\[
L^V_j = \left\{ x = \frac{x_j + x_{j+1}}{2} \right\}, \quad j = 1, \dots, n - 1
\]
For each sweep line \(L\), we compute its intersection with the polygon boundary:
\[
I(L) = \{ p \in \partial P \cap L \}
\]
A polygon is monotone with respect to a given axis if and only if \( |I(L)| = 2 \) for all \(L\) in that axis. If any sweep line satisfies \( |I(L)| > 2 \), the region is non-monotone in that direction, typically due to concavities or internal holes. Note that by construction, the number of candidate sweep lines is at most \(n + m\), corresponding to the grid columns and rows intersecting \(P\), hence monotonicity test scales linearly with grid size.

In accordance with the Feasibility Criterion~\ref{def:acceptability}, if a polygon violates monotonicity in both directions, we initiate decomposition. To determine the appropriate axis and placement for the cut, we use a cumulative gap severity metric, which quantifies the traversal inefficiency along each direction. For a given sweep line \(L\), let \(\{p_1, p_2, \dots, p_{2r}\}\) denote the ordered intersection points of line sweep \(L\) with the polygon boundary. Then, the total gap severity in that axis is defined as:
\begin{equation}
G = \sum_{L \in \mathcal{L}} \sum_{k=1}^{r-1} \|p_{2k+1} - p_{2k}\|
\label{eq:gapseverity}
\end{equation}

where \(\mathcal{L}\) is the set of candidate sweep lines in that axis. Each inner term measures the distance between two consecutive intersection points that fall outside the ROI. These intervals appear either due to internal holes or from disconnected interior spans that arise from the geometry of the polygon. Since a back-and-forth sweep would be forced to cross these intervals, the cumulative sum reflects the total excess distance a naive traversal would incur in the absence of decomposition.

For each sweep line \(L \in \mathcal{L}\) where \( |I(L)| > 2 \), we define a rectangular cross-section bounded between adjacent coordinates in \(X\) or \(Y\), depending on the axis (Equation~\ref{eq:XYcoords}). These cross-sections identify local areas where sweep continuity is interrupted. We denote them as
\[
\mathcal{B}_H = \left\{ B^H_i = P \cap \{ (x, y) : y_i \le y \le y_{i+1} \} \,\middle|\, |I(L^H_i)| > 2 \right\}
\]
\[
\mathcal{B}_V = \left\{ B^V_j = P \cap \{ (x, y) : x_j \le x \le x_{j+1} \} \,\middle|\, |I(L^V_j)| > 2 \right\}
\]

We compare the cumulative severities \(G_H\) and \(G_V\) (see Eq.~\ref{eq:gapseverity}) and choose the axis with greater total. The cut is then placed through the midpoint of the bounding box of the union of cross-sections along that axis. Specifically, letting \(U_H = \bigcup \mathcal{B}_H\) and \(U_V = \bigcup \mathcal{B}_V\), we define the cut line as:
\begin{equation}
\ell =
\begin{cases}
 y = \tfrac{1}{2} (\min\nolimits_y U_H + \max\nolimits_y U_H) 
& \begin{array}{l}
\text{if } G_H \ge G_V \\
\text{and } \mathcal{B}_H \ne \emptyset
\end{array} \\[8pt]
 x = \tfrac{1}{2} (\min\nolimits_x U_V + \max\nolimits_x U_V) 
& \begin{array}{l}
\text{if } G_V > G_H \\
\text{and } \mathcal{B}_V \ne \emptyset
\end{array}
\end{cases}
\label{eq:cutline}
\end{equation}

The decomposition proceeds recursively. After applying a cut \(\ell\) (Equation~\ref{eq:cutline}), the polygon \(P\) is partitioned into child regions \(\{P_1, P_2\}\), each of which is re-evaluated using the same monotonicity test (Criterion~\ref{def:acceptability}) and gap severity analysis. The process recurses on any subregion that remains non-monotone in both directions. Recursion terminates when all subregions satisfy the Feasibility Criterion~\ref{def:acceptability}.

The purpose of the cut \(\ell\) is to distribute the concentration of concavities across the children, so that each inherits a smaller share of the non-monotone support. This tends to reduce concavity complexity of each constituent and encourages larger sweepable regions while keeping the number of partitions low. However, complex geometries can still lead to over-fragmentation. A common failure case occurs when the troublesome cross-sections are highly skewed or clustered. In such cases, a midpoint cut may intersect a narrow indentation and produce thin boundary slivers. To recover from such scenarios, we apply a post-processing merge. Let \(P_1\) and \(P_2\) be two subpolygons produced by the recursion that share a boundary. If their union \(P_1 \cup P_2\) is acceptable under the Uniaxial Feasibility  Criterion~\ref{def:acceptability}, we replace \(\{P_1, P_2\}\) with \(P_1 \cup P_2\) in the partition. This merge operation is applied iteratively to all adjacent pairs until no further merges are possible. Figure~\ref{fig:cpp-1x4-simple} summarizes the main stages of this decomposition workflow on an example polygon.

\subsection{Global Optimization}

The decomposition and merge procedures produce a set of $N$ subpolygon partitions $\{P_1, P_2,\ldots, P_N\}$ that satisfy Criterion~\ref{def:acceptability}. Our goal in this stage is to (i) generate locally efficient candidate coverage paths for each partition, and (ii) globally sequence and connect these into a full, continuous route with minimal overall cost.

For each partition $P_i$, we generate a small set of \textit{candidate} coverage paths $\{\pi_i^{(1)}, \ldots, \pi_i^{(M_i)}\}$, where each $\pi_i^{(j)}$ denotes a \emph{parallel track} traversal that guarantees full coverage of the partition. In this work, we consider up to eight corner-based variants per partition, derived by permuting the sweep direction (horizontal or vertical) and the start corner (bottom-left, bottom-right, top-left, top-right). This gives $M_i=4$ when $P_i$ is monotone with respect to a single axis, and $M_i=8$ when it is monotone to both. 

Each candidate $\pi_i^{(j)}$ is represented as a tuple
\[
\pi_{i}^{(j)} = \big(e_i^{(j)},\, x_i^{(j)},\, L_i^{(j)},\, T_i^{(j)}\big)
\]
where $e_i^{(j)}, x_i^{(j)} \in \mathbb{R}^2$ are entry and exit coordinates, $L_i^{(j)}$ is the total Euclidean path length, and $T_i^{(j)}$ is the number of turns.
We define the local cost of a candidate as
\begin{equation}
\tilde{c}_i^{(j)} \;=\; L_i^{(j)} \;+\; \rho\, T_i^{(j)}
\label{eq:local_cost}
\end{equation}
where $\rho>0$ is a lightweight penalty applied per heading change within the sweep pattern. This term serves as a heuristic bias against excessive direction changes, since each turn typically introduces additional translational acceleration and deceleration overhead in practice.

For any pair of candidate paths $\pi_i^{(j)}$ and $\pi_k^{(\ell)}$, we define a connector cost (path that connects two partitions) as the Euclidean distance from the exit of the first to the entry of the second. This assumes a vehicle that may traverse brief segments outside the ROI when necessary. The connector cost is:

\begin{equation}
d\!\left(x_i^{(j)},\, e_k^{(\ell)}\right) \;=\; \left\| x_i^{(j)} - e_k^{(\ell)} \right\|_2
\label{eq:connector_cost}
\end{equation}

The full planning problem is posed as a joint optimization over both the choice of candidate paths and the visitation order of partitions:
\begin{equation}
\min_{\substack{\sigma \in S_N \\ \mathbf{j}\in \prod_{i=1}^N \{1,\ldots,M_i\}}}
\quad
\sum_{t=1}^{N} \tilde{c}_{\sigma_t}^{(j_{\sigma_t})}
\;+\;
\sum_{t=1}^{N-1} d\!\left(x_{\sigma_t}^{(j_{\sigma_t})},\, e_{\sigma_{t+1}}^{(j_{\sigma_{t+1}})}\right),
\label{eq:global_cost}
\end{equation}
where $\sigma=(\sigma_1,\ldots,\sigma_N)$ is a permutation of $\{1,\ldots,N\}$
$\mathbf{j}=(j_1,\ldots,j_N)$ with $j_i\in\{1,\ldots,M_i\}$ selects a candidate for partition $P_i$,  and $t$ indexes the visitation position along $\sigma$ (i.e., $\sigma_t$ is the $t$-th visited partition). This formulation jointly determines which candidate path to use for each partition and in what order to visit the partitions, so as to minimize the total cost. The first term accounts for the coverage cost within each partition, while the second captures the transition cost between consecutive partitions along the selected tour.

This problem is combinatorially complex, as it requires evaluating all possible permutations of partition visitations along with all combinations of candidate paths; a brute-force approach would yield a time complexity of $O(N! \cdot M^N)$, with $M = \max_i M_i \le 8$. Instead, for small to moderate sets of partitions, we recommend a dynamic programming (DP) formulation that mirrors the classical Held-Karp algorithm for the Traveling Salesman Problem (TSP)\cite{held1962dynamic}. 

The core idea is to build up partial solutions incrementally. We define a DP state $F(S, i, j)$ as the minimum cost of visiting a subset $S \subseteq \{1, \dots, N\}$, ending at partition $P_i$, and using candidate $j \in \{1, \dots, M_i\}$ for that final partition. Each state represents a partial tour that visits exactly the elements of $S$ and terminates with a specific path variant of a specific region. The recurrence relation considers all ways to reach $(i, j)$ by extending a smaller subproblem: for each previously visited partition $P_k \in S \setminus \{i\}$ and candidate $\ell \in \{1, \dots, M_k\}$, we compute the total cost as the sum of (i) the optimal cost to reach $(k, \ell)$, (ii) the local cost $\tilde{c}_i^{(j)}$, and (iii) the connector cost from the exit point of $(k, \ell)$ to the entry point of $(i, j)$. Among all such predecessors, we select the minimum. The base case initializes the cost of visiting a single partition with a single candidate path, and the final solution is obtained by evaluating all complete tours over the full set $\{1, \dots, N\}$. We refer the reader to \cite{alarcon2026pugs_code,  held1962dynamic} for further implementation details.

\subsection{Discussion on Time Complexity} 

As discussed in Section~\ref{sec:approach}, each feasibility test (Criterion~\ref{def:acceptability}) examines \(O(n+m)\) candidate sweep lines, each requiring an intersection query against \(\partial P\). Let \(E\) denote the total number of unit grid edges on \(\partial P\) (including hole boundaries) for the current rectilinear polygon. Under a conservative model in which each sweep-line intersection query costs \(O(E)\), one feasibility test costs \(O((n+m)E)\), which is \(O(E^2)\) in the grid-traced setting since \(n+m \le E\). Over recursive cuts, this cost is applied to each subpolygon, hence \(\sum_i O((n_i+m_i)E_i)\), equivalently \(\sum_i O(E_i^2)\). This is \(O(E^2)\) for balanced splits and degrades to \(O(E^3)\) in the worst case under highly unbalanced recursion. The merge stage then performs repeated pairwise merge attempts with feasibility re-evaluation of adjacent candidate unions. If \(\tilde{N}\) is an upper bound on the number of subpolygons present in any merge pass, then each pass requires \(O(\tilde{N}^2)\) pairwise checks, and repeated passes yield a conservative worst-case merge cost of \(O(\tilde{N}^3 E^2)\).  

The complexity depends on the number of unit grid edges on the boundary, so increasing the grid resolution generally increases runtime (through more sweep lines and more boundary edges to process). If the physical shape is preserved and only the scale changes, the final decomposition remains the same. However in practice, grid resolution is often tied to the upstream segmentation/vision pipeline, which can change the extracted polygon shape (e.g., boundary detail, narrow passages, or holes). In that case, the input geometry itself is different, so the decomposition result may also change.

The global optimizer is a Held-Karp dynamic program over the final (post-merge) partitions. Note that since the number of candidates per partition is bounded $M \le 8$, the overall time complexity of this method is $O(N^2 \cdot 2^N)$, which is significantly more efficient than the brute-force approach, and when paired with our low-partition-count decomposition, it remains tractable for the vast majority of polygonal regions encountered in practice. For large partition counts, we recommend a Nearest-Neighbor–style \cite{alarcon2026pugs_code} heuristic that incrementally connects partition–candidate pairs to minimize cumulative sweep and connector cost.

\section{Evaluation}
\label{sec:evaluation}
\begin{figure*}[t]
  \centering

  \newcommand{\leftcolw}{0.49\linewidth}
  \newcommand{\rightcolw}{0.50\linewidth}
  \newcommand{\leftnudge}{0ex}     
  \newcommand{\rightnudge}{0ex} 

  \begin{minipage}[t]{\leftcolw}
    \vspace*{\leftnudge}%
    \centering
    \includegraphics[width=\linewidth]{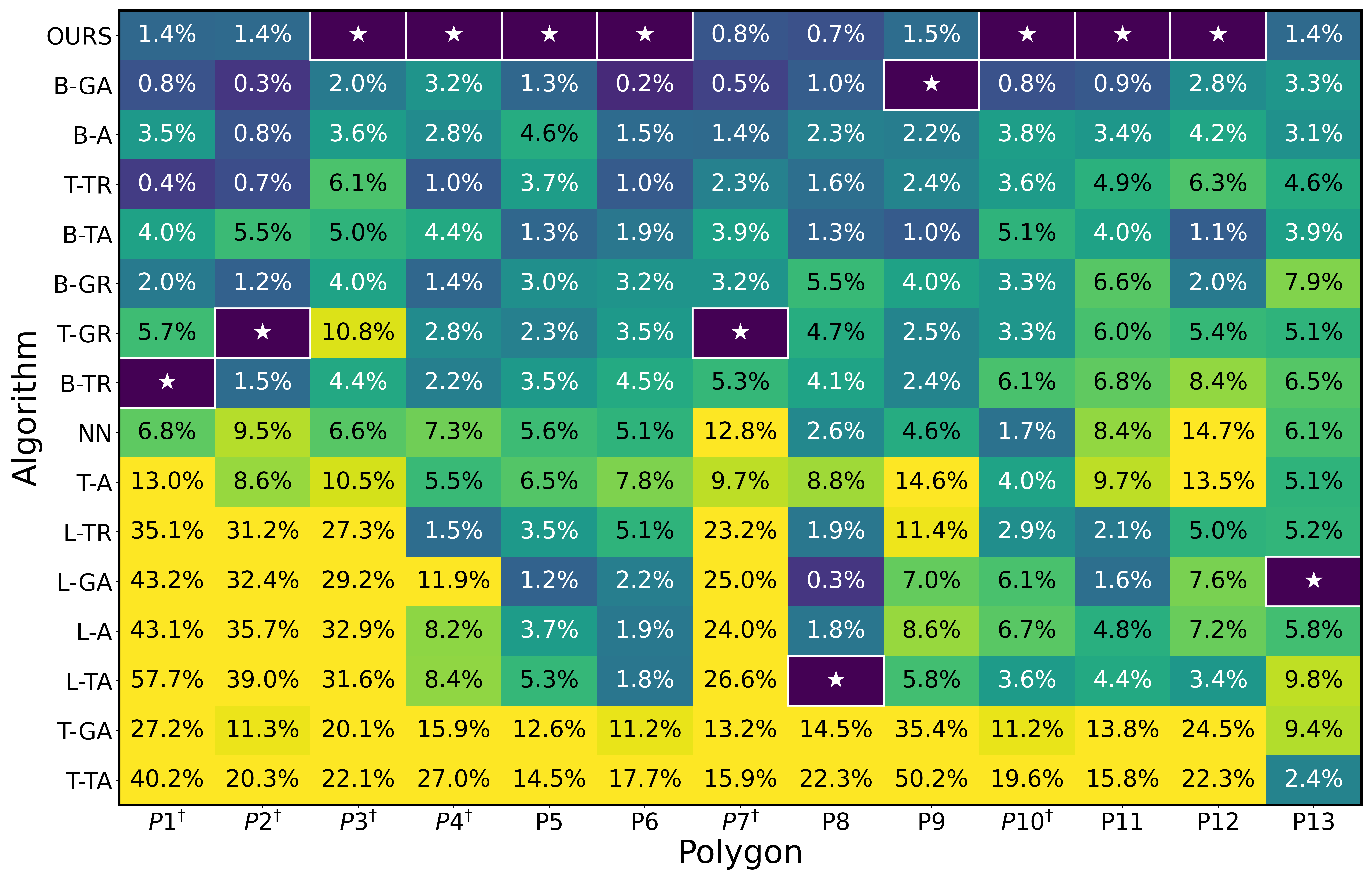}\par\vspace{0.8ex}
    \includegraphics[width=\linewidth]{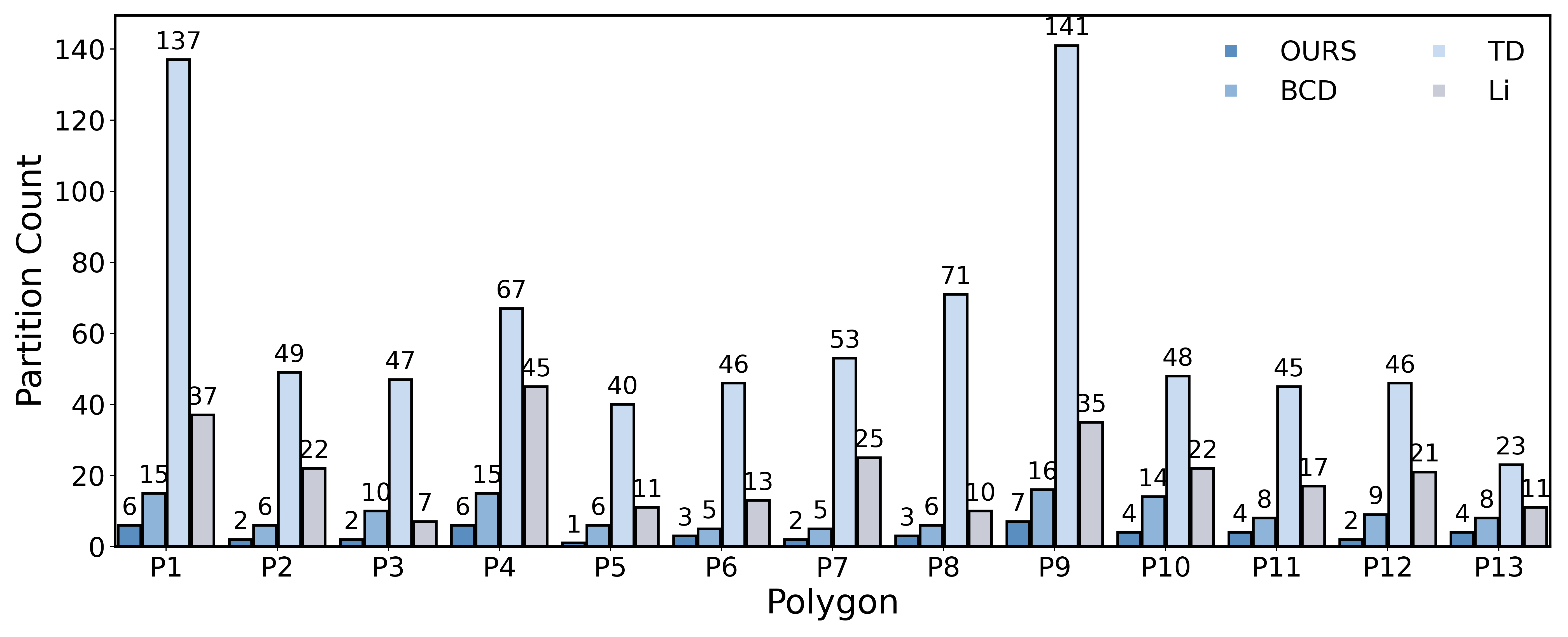}
  \end{minipage}
  \hfill
  \begin{minipage}[t]{\rightcolw}
    \vspace*{\rightnudge}%
    \centering
    \includegraphics[width=\linewidth]{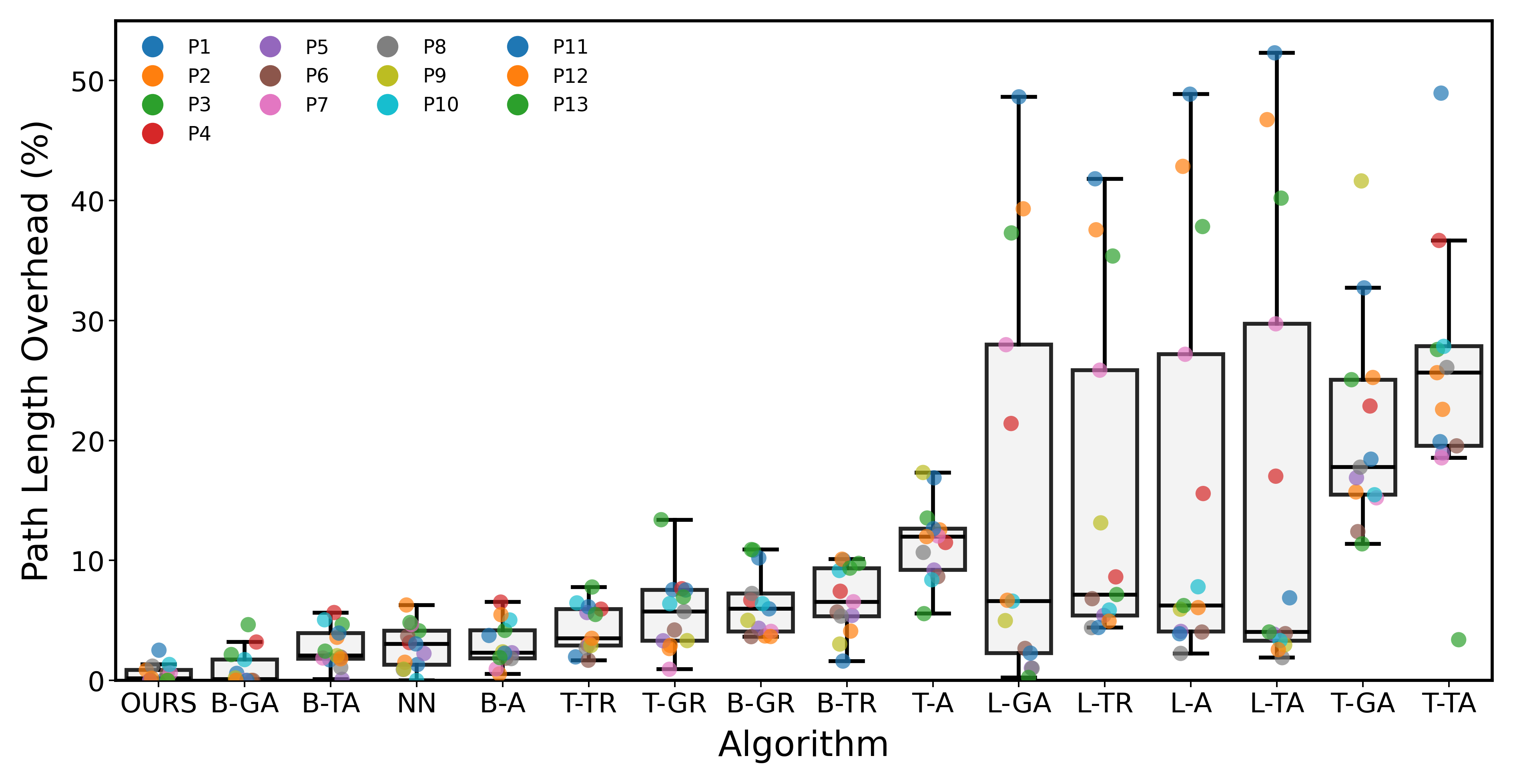}\par\vspace{0.8ex}
    \includegraphics[width=\linewidth]{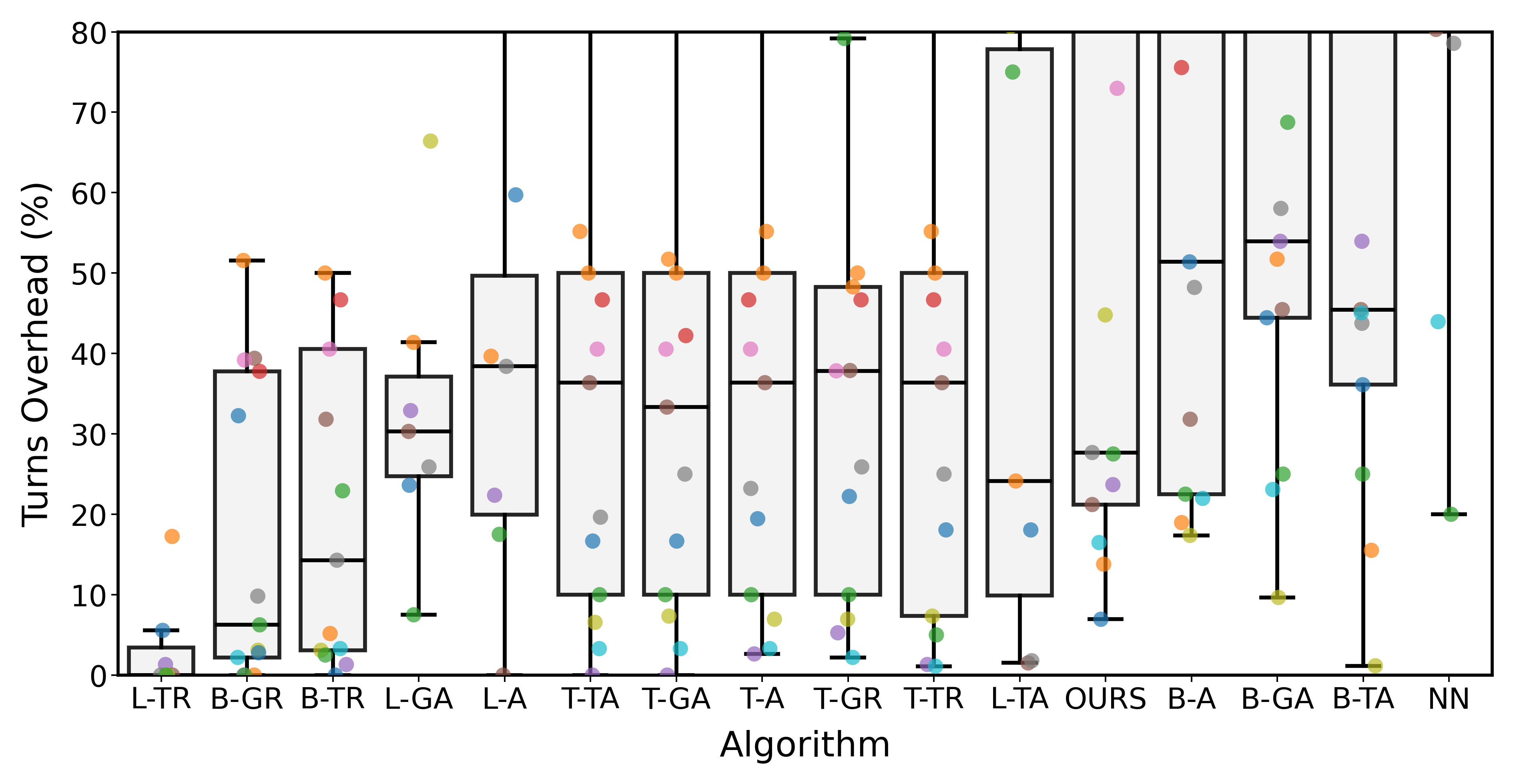}
  \end{minipage}

    \caption{Performance overview across 13 benchmark polygons.
    \textbf{Top-left:} Traversal-time overhead heatmap (\%), sorted by average overhead (darker is better); * marks the best-performing algorithm per polygon. Polygons marked with \({\dagger}\) contain interior holes, Li cannot handle holes, so we fill them before planning with Li. \textbf{Bottom-left:} Partition counts per polygon for each decomposition method (OURS, BCD, TD, LI).
    \textbf{Top-right:} Path length overhead distributions; \textbf{Bottom-right:} Turn overhead distributions.
    Box plots are sorted by mean overhead.
    Algorithms include OURS (proposed), BCD (boustrophedon), T (trapezoidal), L (Li et al.), GA (genetic algorithm), ACO (ant colony optimization), TA (Threshold Accepting), TR (Torres et al.), GR (Google OR-Tools), and NN (nearest neighbor).}
  \label{fig:results-data}
  \vspace{-8pt}
\end{figure*}

\subsection{Experimental Setup \& Baselines}
\label{sec:eval_setup}

We evaluate our method on a benchmark of 13 irregular polygonal ROIs with varied scale, aspect ratio, concavity structure, and internal holes, designed to resemble segmented landcover features (e.g., woodlands, water bodies, and clustered buildings) with geometrical complexities that pose nontrivial challenges to modern CPP pipelines. We construct 15 CPP pipelines by pairing one decomposition method with one global optimizer (i.e., the pairwise combinations considered in this study). The decomposition module outputs a partition set, and the global optimizer determines the global coverage trajectory over those partitions, including candidate sweep selection and visitation order.

We use three decomposition baselines: Trapezoidal Decomposition (TD) \cite{SEIDEL199151}, Boustrophedon Cellular Decomposition (BCD) \cite{choset2000coverage}, and Li's vertex-edge minimum-width decomposition \cite{LI2011876}. TD and BCD are implemented using Fields2Cover \cite{Fields2Cover}, and Li's method is implemented using MathWorks \texttt{coverageDecomposition} \cite{MathWorks_coverageDecomposition}. Li's method, inspired by Huang's minimal-altitude criterion, is a key baseline because it captures the central trade-off in width-based decomposition, namely reduced within-partition turn cost at the expense of increased partition count (i.e., overfragmentation).

For global routing/selection, we use Torres et al.'s heuristic \cite{TORRES2016441} (MathWorks \texttt{uavCoveragePlanner} \cite{MathWorks_UAVCoveragePlanner}), Google OR-Tools \cite{google-ortools} (the default TSP backend used in Fields2Cover), and three reimplemented metaheuristics following the partition-order/sweep-candidate encoding of Jimenez et al. \cite{jimenez2007optimal} with respective hyperparameters listed in Table~\ref{tab:meta-params}: GA, ACO, and Threshold Accepting (TA). As an additional non-decomposition baseline, we include a nearest-neighbor (NN) planner that greedily connects raw sweep waypoints from the original track set. 

Each pipeline outputs a stitched waypoint trajectory that guarantees full coverage. To estimate travel time under realistic motion constraints, we use Ruckig~\cite{berscheid2021jerklimitedrealtimetrajectorygeneration,Ruckig} for offline time parameterization of the stitched waypoint trajectory (i.e., waypoint-to-waypoint line segments). Under our holonomic UAV model (Sec.~\ref{sec:problem-formulation}), omnidirectional planar motion eliminates the need for heading reorientation and curvature-constrained travel between consecutive segments. Hence, traversal time is determined by the translational motion along each segment together with the deceleration/acceleration needed at transitions between consecutive segments. Our Ruckig-based model captures this behavior directly by time-parameterizing each segment under bounded velocity, acceleration, and jerk. The resulting trajectory traversal time serves as our primary performance metric. We additionally report total path length, partition count per decomposition strategy, and number of turns. All metrics are reported as relative overheads with respect to the best value achieved for each polygon, computed as $(M - M_{\min})/M_{\min}$. This normalization enables fair comparison across polygons with different scales and geometric complexity.

\subsection{Analysis \& Discussion of Results}
 
The experimental results in Figure~\ref{fig:results-data} and Table~\ref{tab:aggregate_sumarry} suggest that the decomposition strategy is the primary driver of end-to-end performance. In general, methods that minimize the number of partitions achieve lower time-completion and path-length overheads, with more consistent performance across different polygon shapes. For instance, our proposed method produces the fewest partitions across all polygons, typically in the single digits or low teens (Bar plot in Fig.~\ref{fig:results-data}). This directly contributes to its strong performance, achieving a mean path-length overhead of 0.56\%, seven wins, and eleven top-three finishes. 

This trend extends to other decomposition strategies. BCD produces more partitions than our method but avoids excessive over-fragmentation seen with TD and Li. When paired with GA, BCD becomes the second-best performer, with 1.32\% mean overhead, and nine top-three finishes (Table~\ref{tab:aggregate_sumarry}). In contrast, pipelines based on TD and Li show much weaker performance. The lowest seven ranks in both path-length and execution-time overhead are occupied by TD- or Li-based configurations.  

These results suggest that optimizer choice is typically secondary to decomposition quality. Within each decomposition family, changing the optimizer shifts performance by only 2 to 3 percentage points. For example, BCD ranges from 1.32\% with GA to 4.29\% with Torres. Likewise, Li spans 11.95\% to 15.20\% across optimizers, while TD has more extreme variance, reaching up to 22.33\% in its worst configuration (Table~\ref{tab:aggregate_sumarry}). These patterns suggest the optimizer's limited ability to mitigate poor decomposition. Once decomposition defines the number and geometry of regions, sequencing can only offer marginal gains, but they cannot compensate for the overhead imposed by excessive partition counts or poorly aligned partitions.
\begin{table}[t]
  \centering
  \small
  \caption{Algorithm Performance Summary (Sorted via Lowest Traversal-Time Overhead)}
  \label{tab:aggregate_sumarry}

  \begingroup
  \setlength{\tabcolsep}{4pt}

  \begin{tabular}{l r r r r r}
    \toprule
    \textbf{Alg.} & $\boldsymbol{\mu_T}^{\downarrow}$ (\%) & $\boldsymbol{\mu_L}^{\downarrow}$ (\%) & $\boldsymbol{\mu_K}^{\downarrow}$ (\%) & \textbf{Win}$^{\uparrow}$ & \textbf{Top3}$^{\uparrow}$ \\
    \midrule
    OURS     & 0.56  & 0.56  & 63.88  & 7 & 11 \\
    BCD-GA   & 1.32  & 0.97  & 69.27  & 1 & 9  \\
    BCD-ACO  & 2.85  & 3.00  & 67.17  & 0 & 1  \\
    TD-TR    & 2.97  & 4.33  & 38.15  & 0 & 4  \\
    BCD-TA   & 3.26  & 2.76  & 82.38  & 0 & 3  \\
    BCD-GR   & 3.63  & 6.36  & 17.26  & 0 & 2  \\
    TD-GR    & 4.00  & 5.58  & 37.95  & 2 & 2  \\
    BCD-TR   & 4.29  & 6.74  & 26.10  & 1 & 1  \\
    NN       & 7.06  & 2.82  & 160.40 & 0 & 1  \\
    TD-ACO   & 9.02  & 11.61 & 37.86  & 0 & 0  \\
    LI-TR    & 11.95 & 15.49 & 1.85   & 0 & 0  \\
    LI-GA    & 12.91 & 15.40 & 27.86  & 1 & 4  \\
    LI-ACO   & 14.18 & 16.35 & 35.57  & 0 & 0  \\
    LI-TA    & 15.20 & 16.57 & 63.29  & 1 & 1  \\
    TD-GA    & 16.95 & 20.83 & 37.53  & 0 & 0  \\
    TD-TA    & 22.33 & 27.19 & 37.21  & 0 & 1  \\
    \bottomrule
  \end{tabular}
  \endgroup

  \vspace{0.25em}
  \footnotesize
  \begin{flushleft}
    \textbf{Notation} $\mu_T$: mean traversal-time overhead relative to the per-polygon best; 
    $\mu_L$: mean path-length overhead; 
    $\mu_K$: mean turns overhead; 
    \textbf{Win}: number of first-place finishes; 
    \textbf{Top3}: number of top-three finishes. 
    Arrows: $\uparrow$ higher is better, $\downarrow$ lower is better.
  \end{flushleft}
  \vspace{-12pt}
\end{table}

\begin{figure*}[t]
  \centering
  \newcommand{\panelh}{1.8in}
  \begin{subfigure}[b]{0.24\textwidth}\centering
    \includegraphics[height=\panelh]{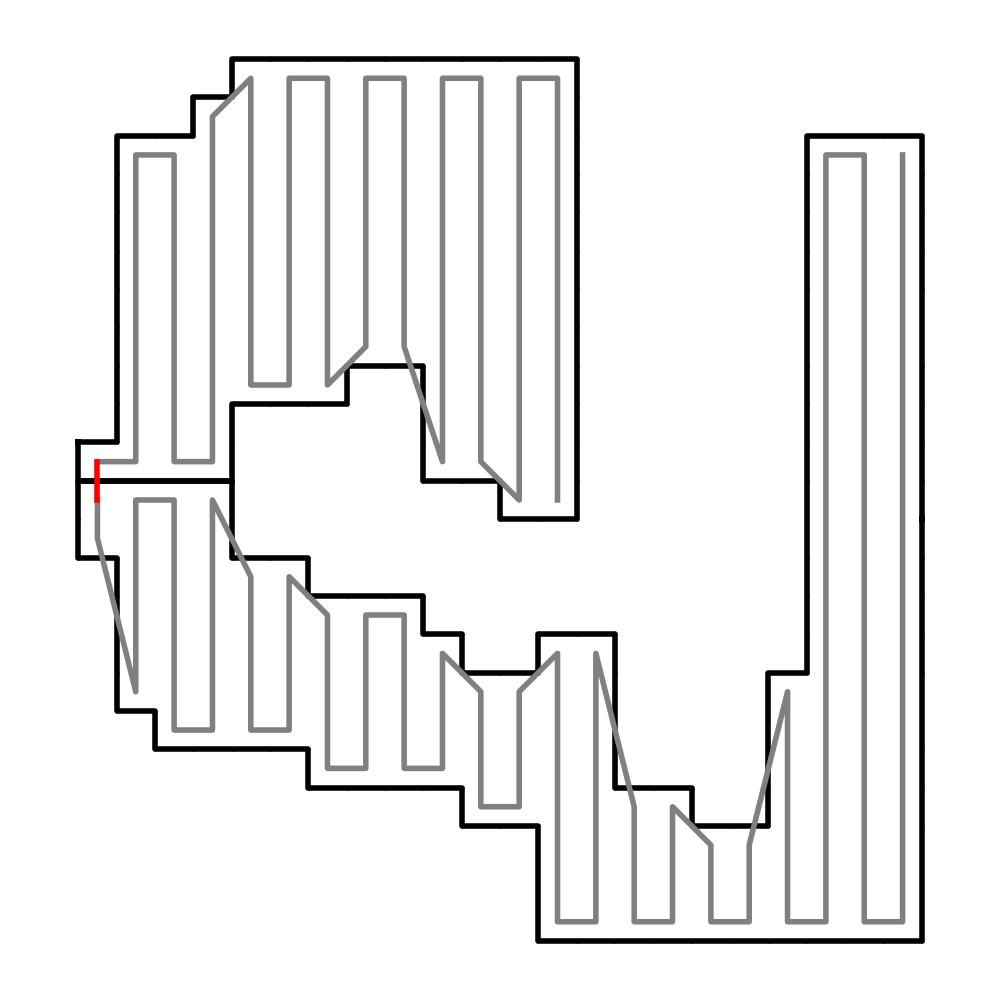}
    \caption{OURS}
  \end{subfigure}\hfill
  \begin{subfigure}[b]{0.24\textwidth}\centering
    \includegraphics[height=\panelh]{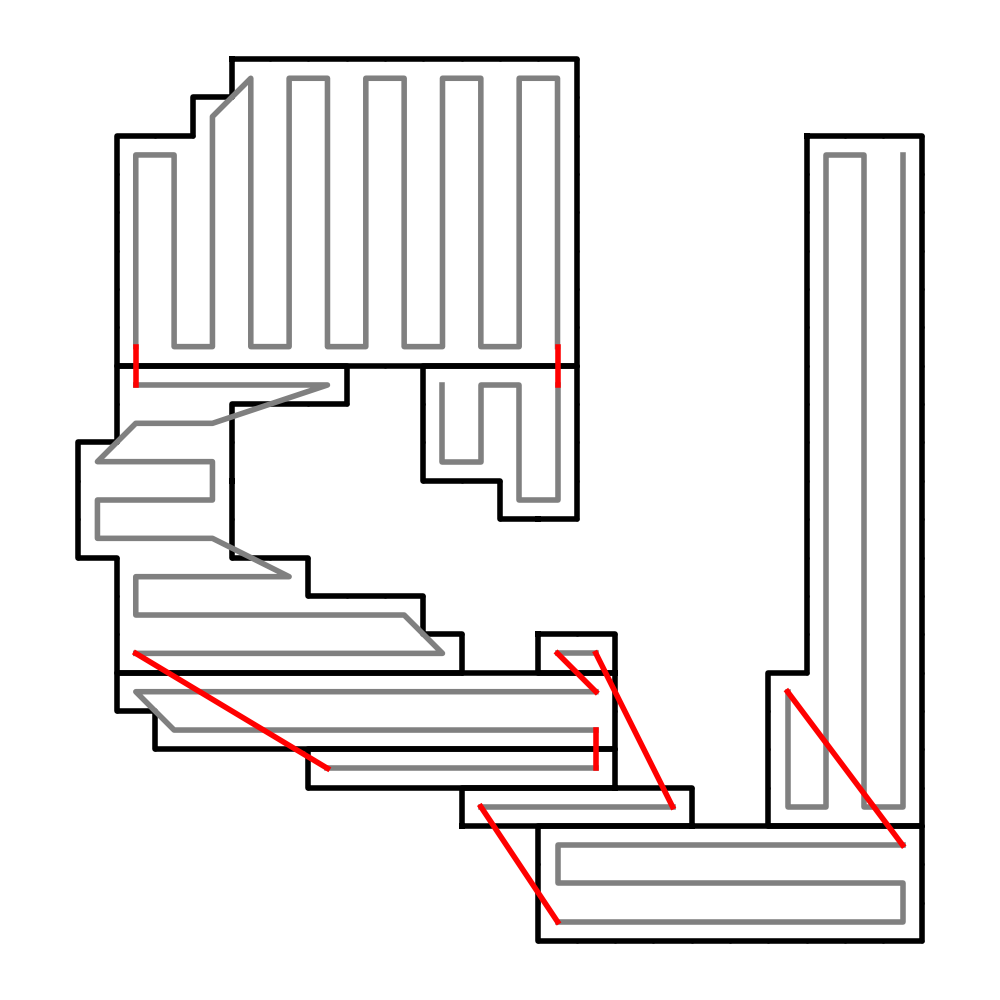}
    \caption{BCD-DSSA}
  \end{subfigure}\hfill
  \begin{subfigure}[b]{0.24\textwidth}\centering
    \includegraphics[height=\panelh]{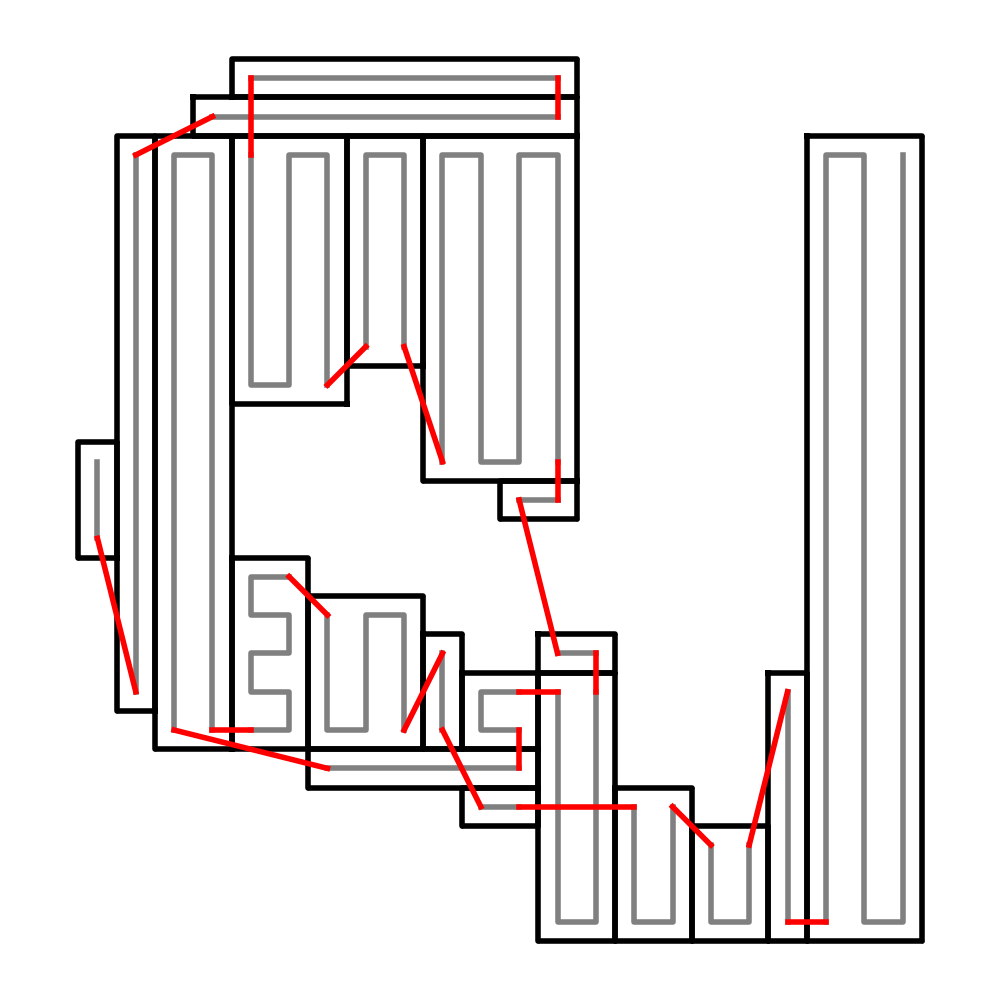}
    \caption{LI-DSSA}
  \end{subfigure}\hfill
  \begin{subfigure}[b]{0.24\textwidth}\centering
    \includegraphics[height=\panelh]{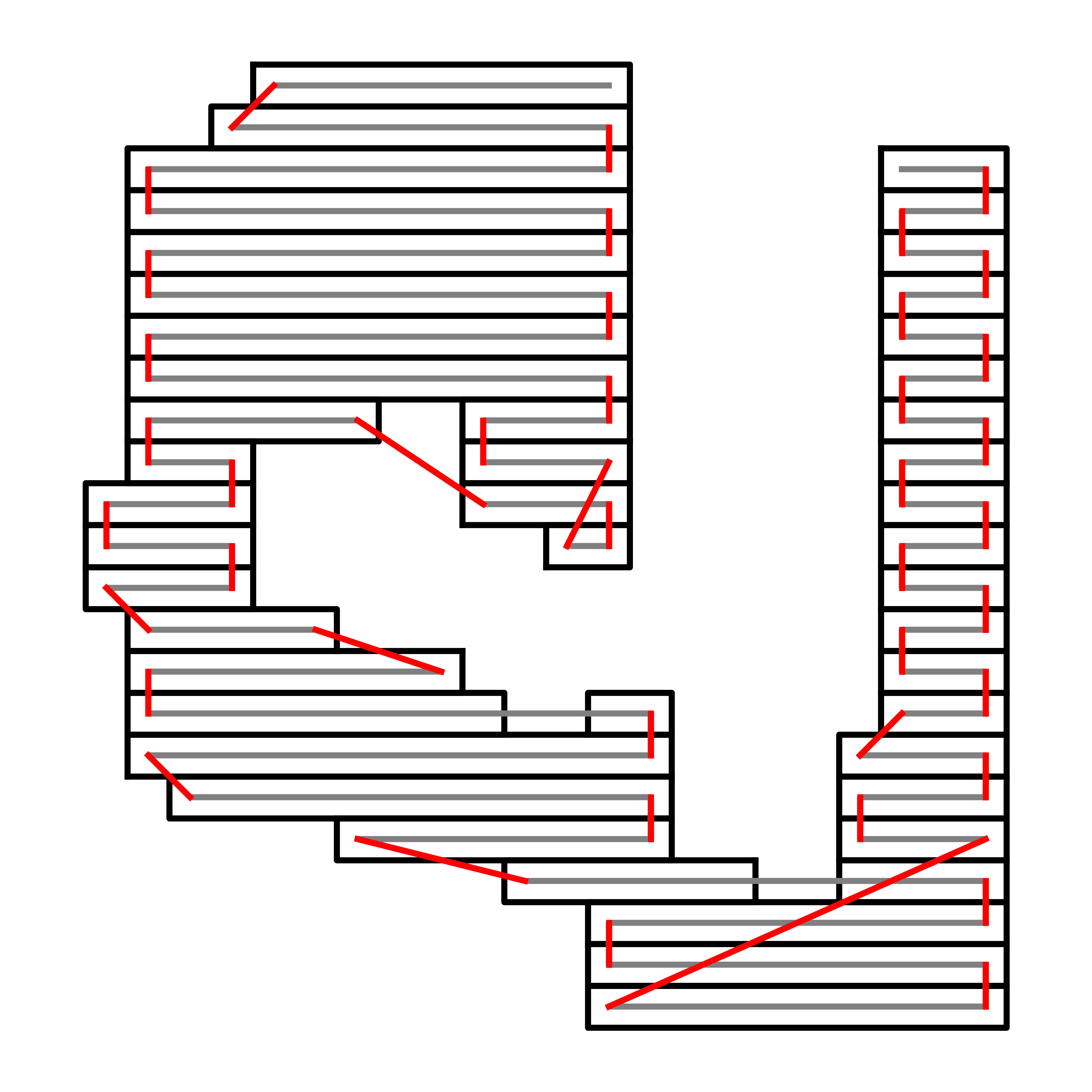}
    \caption{TD-GR}
  \end{subfigure}
  \caption{Final stitched coverage on polygon P12 for OURS and three top-performing baselines, one from each decomposition family. Red lines indicate inter-partition connectors.}
  \label{fig:p12-1x4-final-paths}
  \vspace{-8pt}
\end{figure*}

\begin{table}[t]
  \centering
  \caption{Metaheuristics Parameter Settings}
  \label{tab:meta-params}

  \setlength{\tabcolsep}{2pt}

  \begin{tabular}{ l l r }
    \toprule
    \textbf{Algorithm} & \textbf{Parameter(s)} & \textbf{Value(s)} \\
    \midrule

    \multirow{2}{*}{\textbf{TA}}
      & $\lambda_{\text{turns}},\,K,\,p_{\text{choice}}$ & $0.15,\;20000,\;0.45$ \\
      & $t_0,\,t_1$                                       & $2.0,\;0.02$ \\
    \midrule

    \multirow{3}{*}{\textbf{ACO}}
      & $\lambda_{\text{turns}},\,K,\,w_{\text{elite}},\,m$ & $0.15,\;200,\;2,\;\max(10,\,2\cdot P)$ \\
      & $\alpha_{\text{order}},\,\alpha_{\text{choice}},\,\beta,\,\rho,\,q$ & $1.0,\;1.0,\;2.0,\;0.10,\;1.0$ \\
      & $\tau_0,\,\tau_{\min},\,\tau_{\max}$                 & $1.0,\;10^{-6},\;10^{6}$ \\
    \midrule

    \multirow{2}{*}{\textbf{GA}}
      & $\lambda_{\text{turns}},\,N,\,G,\,f_{\text{elite}},\,k_{\text{tourn}}$ & $0.15,\;450,\;350,\;0.05,\;4$ \\
      & $p_{\text{mut,order}},\,p_{\text{mut,choice}}$                          & $0.30,\;0.40$ \\
    \bottomrule
  \end{tabular}

  \vspace{0.25em}
  \footnotesize
  \begin{flushleft}
    \textbf{Notation} $K$ iterations; $P$ number of partitions; 
    $m=\max(10,\,2\cdot P)$ ants; 
    $\alpha_{\text{order}},\alpha_{\text{choice}}$ pheromone exponents; 
    $\beta$ heuristic exponent; $\rho$ evaporation rate; $q$ deposit scale; 
    $w_{\text{elite}}$ elite deposit weight; 
    $\tau_0,\tau_{\min},\tau_{\max}$ pheromone levels; 
    $N$ GA population size; $G$ generations; $f_{\text{elite}}$ elite fraction; 
    $k_{\text{tourn}}$ tournament size; 
    $p_{\text{mut,order}},p_{\text{mut,choice}}$ mutation probabilities; 
    $\lambda_{\text{turns}}$ turns weight.
  \end{flushleft}
  \vspace{-16pt}
\end{table}

 Figure~\ref{fig:results-data} (turn-overhead box plot) highlights a key limitation of decomposition of width-driven decomposition strategies. While such approaches have been shown to lower turning costs  (as discussed in Section \ref{sec:related-work}), their utility diminishes in the holonomic setting. Specifically, width-based decomposition often increases the total number of partitions, creating a competing objective between within-region turn cost and global path compactness. This trade-off is particularly unfavorable for holonomic UAVs, which can reorient in place with minimal time or energy penalty. The resulting proliferation of narrow partitions leads to longer inter-region connectors and more frequent entry-exit transitions, which inflate global trajectory length and execution time. Moreover, excessive partition counts exacerbate planning complexity. Optimizers that rely on combinatorial reasoning—such as metaheuristics or TSP-style solvers—scale poorly with partition number. As partition count grows, these optimizers either become intractable or must resort to early convergence and local minima, degrading overall performance. Consequently, decomposition strategies that emphasize width minimization at the expense of region count tend to underperform in our context. This effect is evident in the relatively high overheads of width-driven methods like Li, compared to OURS and BCD, which favor structural simplicity and lower partition counts.
 
The Nearest-Neighbor baseline (7.06\% time overhead) further confirms that structured decomposition remains essential even for holonomic vehicles, as greedy point-to-point traversal over raw sweep waypoints produces disorganized and redundant coverage sequences.

These results show that effective coverage planning over irregular polygons hinges on balancing geometric flexibility with partition simplicity. By deferring sweep direction selection and replacing width-based splitting with a cumulative gap severity metric, our method preserves larger, more sweep-efficient regions. This reduces both path length and execution time without incurring the penalties of over-fragmentation.

\section{Conclusion}

This work introduced a decomposition and global optimization strategy tailored to the challenges of holonomic UAV-based CPP in highly irregular environments, such as those found in emergency response missions. We depart from conventional width-based approaches that dominate the CPP literature, and instead we propose a viable alternative built on a uniaxial feasibility criterion and a cut selection process guided by cumulative gap severity. The goal is to distribute clusters of concavity across fewer, structurally sweepable subregions to minimize partition count while preserving parallel-track feasibility. Furthermore, we presented a global optimizer that enumerates sweep candidates for each partition and jointly selects region paths and transition connectors to minimize path length, turn count, and connector cost.

We benchmarked our method against fifteen CPP pipelines comprising established decompositions and state-of-the-art optimizers. Generally, results show that decomposition quality is the dominant factor influencing performance. Methods producing fewer, larger partitions yield shorter execution times and lower path overheads. This advantage stems from reduced inter-partition transitions and fewer axis discontinuities. Our method achieved the lowest mean time and path overheads while demonstrating the most consistent performance across a diverse set of complex polygons. In emergency response applications, the compounding effect of small inefficiencies is far from negligible. Real-world missions often require coverage of dozens or hundreds of disjoint ROIs, where even modest overheads accumulate into significant delays that can impact mission outcomes. Code and benchmark data are publicly available in \cite{alarcon2026pugs_code}.

\bibliographystyle{IEEEtran}
\bibliography{_bibby}

\end{document}